\documentclass{article}

     \usepackage[preprint]{neurips_2021}

\usepackage[utf8]{inputenc} 
\usepackage[T1]{fontenc}    
\usepackage{hyperref}       
\usepackage{url}            
\usepackage{booktabs}       
\usepackage{amsfonts}       
\usepackage{nicefrac}       
\usepackage{microtype}      
\usepackage{xcolor}         
\usepackage{amsmath}
\usepackage{tikz} 
\usepackage{multirow}

\newcommand\mcc[1]{\multicolumn{2}{c}{#1}}

\title{Multi-Agent Algorithmic Recourse}

\author{%
  Andrew O'Brien\\
  Department of Computer Science\\
  Drexel University\\
  Philadelphia, PA 19104 \\
  \texttt{ao543@drexel.edu} \\
   \And
   Edward Kim \\
   Department of Computer Science \\
   Drexel University\\
   Philadelphia, PA 19104 \\
   \texttt{ek826@drexel.edu} \\
}

\begin{document}

\maketitle

\begin{abstract}
	The recent adoption of machine learning as a tool in real world decision making has spurred interest in understanding how these decisions are being made. Counterfactual Explanations are a popular interpretable machine learning technique that aims to understand how a machine learning model would behave if given alternative inputs. Many explanations attempt to go further and recommend actions an individual could take to obtain a more desirable output from the model. These recommendations are known as algorithmic recourse. Past work has largely focused on the effect algorithmic recourse has on a single agent. In this work, we show that when the assumption of a single agent environment is relaxed, current approaches to algorithmic recourse fail to guarantee certain ethically desirable properties. Instead, we propose a new game theory inspired framework for providing algorithmic recourse in a multi-agent environment that does guarantee these properties.
\end{abstract}

\section{Introduction}

Machine learning techniques are increasingly being used to make consequential decisions in the real world \citep{KonigCausal2021}. At the same time, research has  shown machine learning algorithms are capable of learning unethical biases \citep{CaliskanSemantics2017}.  These two trends have lead researchers to begin to examine machine learning techniques that are intelligible to the human decision makers that are using them, so they can ensure decisions are made in a fair way. 

Counterfactual explanations are a popular approach to intelligible machine learning \citep{VermaCounter2020}. The general approach to these explanations is coming up with a systematic way to determine how a machine learning algorithm would have behaved given a different input.  A virtue of this type of explanation is that it allows people who have received a negative evaluation by an algorithm to understand what a successful applicant would look like. 

Algorithmic recourse is an extension of counterfactual explanations that seeks to make action recommendations that would allow an applicant to improve their negative evaluation. One approach to algorithmic recourse is using nearest counterfactual explanations, which involves finding the closest example that would have produced the desirable outcome and recommending making changes to the feature vector to make it resemble the nearest counterfactual example. This is often set up as the  optimization problem (1) below \citep{KarimiRecourse2021}.

\begin{align}
&\delta^{*} \: \epsilon \: \underset{\delta}{ \text{argmin}} \: cost({\delta}; x^F) \\
& s.t. \nonumber \\
& h(x^{CFE}) > t, \nonumber \\
& x^{CFE} = x^{F} + \delta, \nonumber \\
& x^{CFE} \: \epsilon \: P, \nonumber \\
& \delta \: \epsilon \: F, \nonumber \\
\nonumber
\end{align}

The intuition behind the optimization is that we are tying to recommend the lowest cost action, $\delta^{*}$,  that if it were taken would cause the agent's current feature values to shift to a point where the outcome of the machine learning model would be beyond some threshold, t. The cost function for an agent taking an action $\delta$ while at a current feature value $x^F$ is given by $cost({\delta}; x^F)$. The value of the agent's features after taking  action $\delta^{*}$ is denoted by the counterfactual explanation feature vector,  $x^{CFE}$. The value of the machine learning model at some feature vector $x$ is given by $h(x)$. An agent often has only a limited number of actions it can take. For example, it may not be feasible for a loan applicant to increase their savings by more than $10\%$ to change a model's outcome from likely default to unlikely default. The set of feasible actions are denoted by the set $F$. $P$ is a related set that denotes the set of all feature vectors that are actually possible for an agent to achieve. For example, this set will not contain any feature vectors whereby the agent has a different height.

\section{Single vs multi-agent recourse}

Frequently implicit in the literature on nearest counterfactual explanations and algorithmic recourse is the assumption that we only need to take into account the effect an agent's actions have on their own model evaluation \citep{KarimiSurvey2020}. In the real world, a benevolent decision making entity would likely be concerned with the effects of an agent's actions on all of the agents it's advising. In strategic environments, an individual's utility often depends on its own actions and the actions of other agents. Therefore, recommended actions that are  are optimal from an individual agent's perspective may not be optimal when considered from a multi-agent perspective. 

The following example, based on the classic strategic game the prisoner's dilemma, is illustrative of the problems that can arise when using recourse techniques based on a single-agent perspective. Assume an entity is advising two agents on how to reduce their current prison sentence. Each agent has the option to collaborate with the police and betray the other agent or remain silent and refuse to collaborate. The sentence reductions for each agent are a function of both their own action and the action of the other agent and are given by table 1.  Assume the entity has  models $h_{1}$ and $h_{2}$ that perfectly predict the payoffs given in the table below for agents one and two respectively. 
    \begin{table}[h!]
    \begin{center}
    \caption{\label{tab:table 1} Payoff matrix for prisoner's dilemma example.}
    \renewcommand\arraystretch{1.3}
\begin{tabular}{cccc}

\mcc{}          &   \mcc{Agent 2}         \\
    \cline{2-4}
    &           &   Betray (1)   &   Silent (0)       \\
    \cline{2-4}
\multirow{2}{*}{Agent 1}
    &   Betray (1) &   3.5,3.5     &   10,1     \\
    \cline{2-4}
    &   Silent (0)   &   1,10     &   5,5     \\
    \cline{2-4}

\end{tabular}

    \end{center}

    \end{table}

Assume that agent one comes to the entity and notes it's current sentence reduction is given by $h_{1}(x^{F} = [0, 1]) = 1$ years. $P = \{[0, 0], [1, 0], [0, 1], [1, 1]\}$ and $F = \{[1, 0], [0,0]\}$. In this example, the optimal recourse according to the nearest counterfactual explanation is $\delta^{*} = [1, 0]$. The new predicted payoff for agent one is $h_{1}([1, 1]) = 3.5$. The goal of recourse for that agent has been achieved. There are two additional facts to note. $h_{2}$ has gone from 10 to 3.5 on the new input $x^
{CFE}  = [1,1]$. Furthermore, $h_{1}(x^{CFE}) + h_{2}(x^
{CFE}) = 7 < h_{1}(x^{F}) + h_{2}(x^{F}) = 11$. The algorithmic recourse recommendation has made agent two worse off and decreased the sum of benefits to both agents. 

Now consider the alternate case where agent one comes to the entity and notes it's current sentence reduction is given by $h_{1}(x^{F} = [0, 0]) = 5$ The recourse recommendation that improves agent one's outcome is
$\delta^{*} = [1, 0]$ As before, this recommendation makes agent two worse off. Unlike the previous case, the amount it makes agent agent one better off is greater than the amount it makes agent two worse off, so the group is better off.

The above examples illustrates inherent trade-offs that must be considered when making recourse recommendations in strategic situations where an agent's outcome depends on her actions and those of other agents. An ethically ideal recommended action would improve the prediction of the agent being advised (principle agent), not worsen the prediction for any other agent, and increase the sum of the predictions for all of the agents. The latter two properties are known in game theory as Pareto efficiency and social welfare efficiency \citep{TadelisGame2013}. When considering the single agent environment, any action that achieves one must trivially satisfy all three. However, when there are at least two agents, an action that achieves one goal can fail to achieve one or both of the other two. 

Our work is not the first to examine recourse from a multi-agent perspective. \citet{RawalBeyond2020} propose a framework for multi-agent recourse but do not incorporate causal relationships between features when making recommendations. Furthermore, although they take into account that recourse recommendations might perform differently for different subgroups , they don't address the idea that an agent acting on a recommendation might affect another agent's model outcome. 

\section{Counterfactual identification}

In order to make a multi-agent recourse recommendation, the effect of the recommendation on the predictions for all the agents must be known. Causal and counterfactual knowledge of this kind can be encoded as a structural causal model M and a corresponding causal graph $G_{M}$\citep{PearlCause2009}. M = <F, X, U> is a three tuple of endogenous variables X, exogenous variables U, and a set of structural equations $F = \{f_{1},..., f_{|X|}\}$ that determine how values are assigned to the endogenous variables. In the directed causal graph $G_{M}$, there is a node for every exogenous and endogenous variable. An edge connects a node $n_{i}$ to $n_{j}$ if and only if the variable corresponding to no $n_{i}$ appears as an argument in the function $f_{j}$ which assigns values to the variable corresponding to $n_{j}$. 

Counterfactual statements can be represented using do-operations of the form $do(X_{i} := k)$. A do-operation transforms M to $M_{A}$ by assigning k to $X_{i}$ instead of assigning it the value of $f_{i}$. The corresponding causal graph $G_{M_{A}}$ is identical to G but with the incoming edges to the node corresponding to $X_{i}$ removed. Recommended actions can be interpreted as a set of do-interventions $A = do(\{X_{i} := a_{i}\}_{i \epsilon I})$. Assuming no hidden confounders and an invertible F, any X can be uniquely determined given any U and vice versa. Hence any structural counterfactual query can be computed in the following way $x^{SCF} = F_{A}( F^{-1}(x^{F}))$ \citep{KarimiRecourse2021}.

The prisoner's dilemma example can be stated as the following structural causal model. $M = <F, X, U>$ where  $X = \{h_{1}, h_{2}\}$, $U = \{x_{1}, x_{2} \}$, $F= \{ f_{1}, f_{2}\}$, and

\[
    f_{i}(x_{i}, x_{j})= 
\begin{cases}
    1              & \text{if} \: x_i = 0, x_j = 1 \\
    3.5              & \text{if} \: x_i = 1, x_j = 1 \\
    5              & \text{if} \: x_i = 0,  x_j = 0 \\
    10              & \text{if} \: x_i = 1,  x_j = 0 \\
\end{cases}
\]

The corresponding causal graph $G_{M}$ is depicted in figure 1.

\begin{center}

\begin{tikzpicture}[main/.style = {draw, circle}, node distance=4cm] 
    \node[main] (1)  {$x_1$}; 
    \node[main] (2) [right of=1] {$x_2$};
    \node[main] (3) [ below of =1] {$h_1$};
    \node[main] (4) [ below of =2] {$h_2$};
    \draw[->] (1) -- (3);
    \draw[->] (1) -- (4);
    \draw[->] (2) -- (3);
    \draw[->] (2) -- (4);
\end{tikzpicture}

\end{center}

\begin{figure}
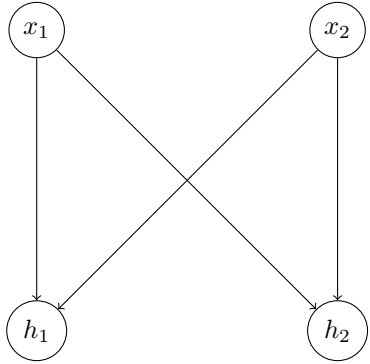

\begin{center}
\caption{\label{fig:figure 1}Figure 1: The causal graph corresponding to the prisoner's dilemma example.}
\end{center}
\end{figure}

Learning causal graphs from data is currently an open area of research \citep{DemlStructure2018}. Although progress has been made in recent years,  the difficulty of learning a causal graph or structural causal model from data is a serious limitation on any approach to recourse, such as ours, that assumes this knowledge.

\section{Problem reformulation}

Below we present three different algorithmic recourse optimization problems: single agent efficient, social welfare efficient, and Pareto efficient. Each problems corresponds to a different desirable ethical property that the recommendation should have.  

\subsection{Single agent efficient recourse}

\citet{KarimiRecourse2021} showed that incorporating causal models was necessary for making accurate recourse commendations in the case of single agent recommendations. They restate the problem of making recourse recommendations as one of making minimal interventions as follows. 

\begin{align}
& a^{*} \: \epsilon \: \underset{a \epsilon A}{ \text{argmin}} \: cost_{i}(a; x^F) \\
& s.t. \nonumber \\
& h_{i}(x^{SCF, a}) \geq t, \nonumber \\
& x^{SCF, a} = F_{a}(F^{-1}(x^F)), \nonumber \\
& x^{SCF, a} \: \epsilon \: P, \nonumber \\
\nonumber
\end{align}

The intuition behind the problem is that we are trying to recommend an action a that minimizes agent i's cost function while meeting a set of constraints that ensure action a changes the feature vector to one that causes the model predicting agent i's outcome, $h_{i}$, to exceed some threshold, t. The new feature vector is the structural counterfactual outcome and denoted $ x^{SCF}$. As before, $P$ is set of plausible actions. This is the correct statement of the problem when the goal is to maximize the outcome for the principle agent. 

\subsection{Social welfare efficient recourse}

In the case that the entity wants to recommend an action that increases the sum of the predictions for the N-agents it advises, the following is the correct optimization problem. 

\begin{align}
& a^{*} \: \epsilon \: \underset{a \epsilon A}{ \text{argmin}} \: cost_{i}(a; x^F) \\
& s.t. \nonumber \\
& \sum_{j=1}^N h_{j}(x^{SCF}) > \sum_{j=1}^N h_{j}(x^{F}) , \nonumber \\
& x^{CFE, a} = F_{a}(F^{-1}(x^F)), \nonumber \\
& x^{CFE, a} \: \epsilon \: P_{i}, \nonumber \\
\nonumber
\end{align}

An alternate variation is one where the sum of the predictions isn't decreasing and first constraint isn't strict. 

\subsection{Pareto efficient Recourse}

In the case that the entity wants to recommend an action that makes none of the N agents the entity advises worse off, the following is the correct optimization problem. 

\begin{align}
    & a^{*} \: \epsilon \: \underset{a \epsilon A}{ \text{argmin}} \: cost_{i}(a; x^F) \\
    & s.t. \nonumber \\
    & \forall j \: \epsilon \: N, h_{j}(x^{SCF}) \geq h_{j}(x^{F}) , \nonumber \\
    & x^{CFE, a} = F_{a}(F^{-1}(x^F)), \nonumber \\
    & x^{CFE} \: \epsilon \: P_{i}, \nonumber \\
    \nonumber
\end{align}

The inequality constraints from 2-4 can be combined. For example, an institution might find it ethically desirable to recommend actions that make agent i strictly better off but make no other agent worse off or the group no worse off. 

\section{Experimental results}

We tested the effects of our multi-agent optimization using experimental data from \citet{BoCoop2005}. B\'o ran an experiment testing players' strategies in a simulation of an iterated prisoner's dilemma with infinite rounds. The iterated prisoner's dilemma is a multi-round variant of the prisoner's dilemma whereby a player's outcome is the sum of their outcomes in the individual rounds of the game.  In his experiment, he split players into a test and control group. The test group, designed to simulate infinite play, would have each game consist of an initial round and then an unknown number of additional rounds where each additional round had a probability $\delta$ of occurring.  $\delta$ was set to 0, 1/2, or 3/4.  The corresponding control group players played a fixed number of games. The fixed number was set to either 1 game if it was a control for $\delta = 0$ , 2 games for $\delta = 1/2$, or 4 games for $\delta = 3/4$, thus ensuring the control and test group played the same number of games on average. Test games with $\delta$ equal to 0 and control games with one round are equivalent to the single round prisoner's dilemma from section 2. 3294 games met one these criterion. Participants were paid an amount proportional to the number of the points they won in the game. Two different pay-off matrices were used during the game and are displayed below. 

\begin{table}[hbt!]
    \begin{center}
    \caption{\label{tab:table 2} First payoff matrix used in experiments from \citet{BoCoop2005}.}
    \renewcommand\arraystretch{1.3}
    \begin{tabular}{cccc}

\mcc{}          &   \mcc{Agent 2}         \\
    \cline{2-4}
    &           &   Betray (1)   &   Silent (0)       \\
    \cline{2-4}
\multirow{2}{*}{Agent 1}
    &   Betray (1) &   35,35     &   100,10     \\
    \cline{2-4}
    &   Silent (0)   &   10,100     &   65,65     \\
    \cline{2-4}

    \end{tabular}

    \end{center}

\end{table}

\begin{table}
\begin{center}
    \caption{\label{tab:table 3} Second payoff matrix used in experiments from \citet{BoCoop2005}.}
    \renewcommand\arraystretch{1.3}
\begin{tabular}{cccc}

\mcc{}          &   \mcc{Agent 2}         \\
    \cline{2-4}
    &           &   Betray (1)   &   Silent (0)       \\
    \cline{2-4}
\multirow{2}{*}{Agent 1}
    &   Betray (1) &   45,45     &   100,10     \\
    \cline{2-4}
    &   Silent (0)   &   10,100     &   75,75     \\
    \cline{2-4}

\end{tabular}

\end{center}

\end{table}

 We examined the effects of recommendations made using single-agent optimization constrains. Of the recommendations made for the 3294 single-round prisoner's dilemma games, 434 recommendations resulted in an improvement for the principle agent as well as both a loss in social welfare and a violation of Pareto efficiency. No recommendations resulted in either social welfare or the opposing player improving. When either Pareto efficient or social welfare efficient constraints were added or just Pareto efficient constraints were considered alone, no recommendations were made at all because no action can improve the principle agent while improving social welfare or not harming the other player. Therefore, optimization setups from our framework would have prevented significant third party harm when compared with single agent recourse recommendations.
 
When recommendations were made with just social welfare constraints, 2860 recommendations resulted in both an increase in social welfare and a decrease in the principle agent's welfare. None of the recommendations resulted in an increase in the principle agent's welfare. This result highlights a pressing ethical dilemma that previous work using the single agent framework obscured, whether or not to advise the principle agent to do something that makes it worse off because it makes the group better off. It is important that the artificial intelligence community develops a consensus on how to handle dilemmas such as this.

\section{Discussion and future work}

As machine learning models become increasingly important to organizational decision making, the need for realistically understanding the effects of those decisions will grow. Furthermore, the demand will grow to ensure machine learning powered decision making takes into account its effects on all stakeholders. Pareto superior and social welfare efficient constraints when making recourse recommendations are a step towards this inclusive style of machine learning based decision making.

This paper provided both theoretical and empirical support for the argument that an inherent trade off exists between making recourse recommendations that are good for the group and recommendations that are good for the principle agent. In the future, comparing how these different approaches to recourse recommendations perform on additional real world data sets could demonstrate concrete benefits to this line of research. 

For example, companies such as Apple, Google, and Waze offer GPS services  make route recommendations to many drivers simultaneously. The time it takes a driver to reach its destination is in part a function of the other drivers on the road. It is established that certain network configurations can result in a prisoner's dilemma whereby individual drivers taking the fastest route makes all of the drivers worse off \citep{NisanAlg2007}. There is evidence that this phenomenon is currently contributing to traffic congestion in the real world \citep{CabannesImpact2018}.  A GPS application making recommendations in a traffic network may have to choose between making recommendations that are single agent optimal, social welfare efficient, or Pareto efficient. To the best of the authors' knowledge, no major GPS company has addressed this problem in the literature or otherwise. Our work can have an immediate impact on real world recourse problems such as this. 

Finally, other ethical constraints in addition to Pareto efficiency and social welfare efficiency should be explored. 

\bibliographystyle{unsrtnat}
\bibliography{references}

\end{document}